\documentclass[10pt,twocolumn,letterpaper]{article}

\usepackage[final]{cvpr}      

\usepackage{graphicx}
\usepackage{amsmath}
\usepackage{amssymb}
\usepackage{amsthm}
\usepackage{booktabs}
\usepackage{mathtools}
\usepackage{algorithm}
\usepackage{algorithmicx}
\usepackage{algpseudocode}

\usepackage[pagebackref,breaklinks,colorlinks]{hyperref}

\usepackage[capitalize]{cleveref}
\crefname{section}{Sec.}{Secs.}
\Crefname{section}{Section}{Sections}
\Crefname{table}{Table}{Tables}
\crefname{table}{Tab.}{Tabs.}

\def\cvprPaperID{} 
\def\confName{}
\def\confYear{}

\begin{document}

\title{SLOSH: Set LOcality Sensitive Hashing \\ via Sliced-Wasserstein Embeddings}

\author{Yuzhe Lu$^*$\\
Computer Science Department,\\
Vanderbilt University,\\
Nashville, TN, 37235\\
{\tt\small yuzhe.lu@vanderbilt.edu}
\and
 Xinran Liu$^*$\\
Computer Science Department,\\
Vanderbilt University\\
 Nashville, TN, 37235\\
{\tt\small xinran.liu@vanderbilt.edu}
\and
Andrea Soltoggio\\
School of Computer Science,\\
Loughborough University,\\
Leicestershire, UK\\
{\tt\small a.soltoggio@lboro.ac.uk}
\and
Soheil Kolouri\\
Computer Science Department,\\
Vanderbilt University,\\
Nashville, TN, 37235\\
{\tt\small soheil.kolouri@vanderbilt.edu}
}
\maketitle

\begin{abstract}
   Learning from set-structured data is an essential problem with many applications in machine learning and computer vision. This paper focuses on non-parametric and data-independent learning from set-structured data using approximate nearest neighbor (ANN) solutions, particularly locality-sensitive hashing. We consider the problem of set retrieval from an input set query. Such retrieval problem requires: 1) an efficient mechanism to calculate the distances/dissimilarities between sets, and 2) an appropriate data structure for fast nearest neighbor search. To that end, we propose Sliced-Wasserstein set embedding as a computationally efficient ``set-2-vector'' mechanism that enables downstream ANN, with theoretical guarantees. The set elements are treated as samples from an unknown underlying distribution, and the Sliced-Wasserstein distance is used to compare sets. We demonstrate the effectiveness of our algorithm, denoted as Set-LOcality Sensitive Hashing (SLOSH), on various set retrieval datasets and compare our proposed embedding with standard set embedding approaches, including Generalized Mean (GeM) embedding/pooling, Featurewise Sort Pooling (FSPool), and Covariance Pooling and show consistent improvement in retrieval results. The code for replicating our results is available here: \href{https://github.com/mint-vu/SLOSH}{https://github.com/mint-vu/SLOSH}.
\end{abstract}

\section{Introduction}
\label{sec:intro}

The nearest neighbor search problem is at the heart of many nonparametric learning approaches in classification, regression, and density estimation, with many applications in machine learning, computer vision, and other related fields \cite{andoni2006near,shakhnarovich2008nearest,andoni2015optimal}.    The exhaustive search solution to the nearest neighbor problem for $N$ given objects (e.g., images, vectors, etc.) requires $N$ evaluation of (dis)similarities (or distances), which could be problematic when: 1) the number of objects, $N$, is large, or 2) (dis)similarity evaluation is expensive.  Approximate Nearest Neighbor (ANN) 
\cite{andoni2018approximate} approaches have been proposed as an efficient alternative for similarity search on massive datasets.  ANN approaches leverage data structures like random projections, e.g., Locality-Sensitive Hashing (LSH) \cite{indyk1998approximate,datar2004locality}, or tree-based structures, e.g., kd-trees \cite{bentley1975multidimensional,wald2006building}, to reduce the complexity of nearest neighbor search. Ideally, ANN approaches must address both these challenges, i.e., decreasing the number of similarity evaluations and reducing the computational complexity of similarity calculations while providing theoretical guarantees on ANN retrievals.

Despite the great strides in developing ANN methods, the majority of the existing approaches are designed for objects living in Hilbert spaces. Recently, however, there has been an increasing interest in set-structured data with many applications in point cloud processing, graph learning, image/video recognition, and object detection, to name a few \cite{zaheer2017deep,lee2019set,wagstaff2019limitations}. Even when the input data itself is not a set, in many applications, the complex input data (e.g., a natural image or a graph) is decomposed into a \emph{set} of more abstract components (e.g., objects or node embeddings). Similarity search for large databases of set-structured data remains an active field of research, with many real-world applications. In this paper, we focus on developing a data-independent ANN method for set-structured data. We leverage insights from computational optimal transport \cite{villani2008optimal,bonnotte2013unidimensional,kolouri2017optimal,peyre2018computational} and propose a novel LSH algorithm, which relies on Sliced-Wasserstein Embeddings and enables efficient set retrieval.  

Defining (dis)similarities for set-structured data comes with unique challenges: i) the sets could have different cardinalities, and ii) the set elements do not necessarily have an inherent ordering. Hence, a similarity measure for set-structured data must handle varied input sizes and should be invariant to permutations, i.e., the (dis)similarity score should not change under any permutation of the input set elements. Generally, the existing approaches for defining similarities between sets rely on the following two strategies. First, solving an assignment problem (via optimization) for finding corresponding elements between two sets and aggregate (dis)similarities between corresponding elements, e.g., using Hungarian algorithm, Wasserstein distances, Chamfer loss, etc. These approaches are at best quadratic and at worst cubic in the set cardinalities.

The second family of approaches rely on embedding the sets into a vector space and leveraging common similarities in the embedded space. The set embedding could be explicit (e.g., deep set networks) \cite{zaheer2017deep,lee2019set} or implicit (e.g., Kernel methods) \cite{jebara2004probability,gretton2006kernel,boiman2008defense,poczos2011nonparametric,poczos2012nonparametric,muandet2012learning,xiong2014learning,kolouri2016sliced}. Also, the embedding process could be data-dependent (i.e., learning based) as in deep set learning approaches, which leverage a composition of permutation-equivariant backbones followed by a permutation-invariant global pooling mechanisms that define a parametric permutation-invariant set embedding into a Hilbert space \cite{zaheer2017deep,lee2019set,Zhang2020FSPool}. Or, it can be data-independent as it is the case for global average/max/sum/covariance pooling, variations of Janossy pooling \cite{murphy2018janossy}, and variations of Wasserstein embedding \cite{kolouri2021wasserstein}, among others.  Recently, there has been a lot of interest in learning-based embeddings using deep neural networks and in particular transformer networks. However, data-independent embedding approaches (e.g., global poolings) have received less attention.



{\bf Contributions.} Our paper focuses on non-parametric learning from set-structured data using data-independent set embeddings. Precisely, we consider the problem where our training data is a set of sets, i.e., $\mathcal{X}=\{X_i | X_i=\{x^i_n\in\mathbb{R}^d\}_{n=0}^{N_i-1}\}_{i=1}^I$, (e.g., set of point clouds), and for a query set $X$ we would like to retrieve the $K$-Nearest Neighbors (KNN) from $\mathcal{X}$. To solve this problem, we require a fast and reliable (dis)similarity measure between sets, and a computationally efficient nearest neighbor search. We propose Sliced-Wasserstein Embedding as a computationally efficient and powerful tool that provides a set-2-vector operator with computational complexity of $\mathcal{O}(LN(d+log(N))$, with sequential processing, $\mathcal{O}(N(d+log(N))$, with parallel processing, where $L$ is of the same order as $d$. Treating sets as empirical distributions, Sliced-Wasserstein Embedding embedds sets in a vector space in which the Euclidean distance between two embedded vectors is equal to the Sliced-Wasserstein distance between their corresponding empirical distributions. Such embedding enables the application of fast ANN approaches, like Locality Sensitive Hashing (LSH), to sets while providing collision probabilities with respect to the Sliced-Wasserstein distance. Finally, we provide extensive numerical results analyzing and comparing our approach with various data-independent embedding methods in the literature. 

\section{Related Work}
\label{sec:related}
{\bf Set embeddings (set-2-vector):} Machine learning on set-structured data is challenging due to: 1) permutation-invariant nature of sets, and 2) having various cardinalities. Hence, any model (parametric or non-parametric) designed for analyzing set-structured data has to be permutation invariant, and allow for inputs of various sizes. Today, a common approach for learning from sets is to use a permutation equivariant parametric function, e.g., fully connected networks \cite{zaheer2017deep} or transformer networks \cite{lee2019set}, composed with a permutation invariant function, i.e., a global pooling, e.g., global average pooling, or pooling by multi-head attention \cite{lee2019set}. One can view this process as embedding a set into a fixed-dimensional representation through a parametric embedding that could be learned using the training data and an objective function (e.g., classification). 

A major unanswered question is regarding non-parametric learning from set-structured data. In other words, what would be a good data-independent set embedding that one can use for generic applications, including K-Nearest Neighbor classification/regression/density estimation? Given that a data-independent global pooling could be viewed as a set-2-vector process, we surveyed the existing set-2-vector mechanisms in the literature. In particular, global average/max/sum and covariance pooling \cite{wang2020deep} could be considered as the simplest such processes. Generalized Mean (GeM) \cite{radenovic2018fine} is another pooling mechanism commonly used in image retrieval applications, which captures higher statistical moments of the underlying distributions. Other notable approaches include VLAD \cite{jegou2010aggregating,arandjelovic2013all}, CroW \cite{kalantidis2016cross}, and FSPool \cite{Zhang2020FSPool}, among others. 

{\bf Locality Sensitive Hashing (LSH):} A LSH function hashes two ``similar'' objects into the same bucket with ``high'' probability, while ensuring that ``dissimilar'' objects will end up in the same bucket with ``low'' probability. Originally presented in \cite{indyk1998approximate} and extended in \cite{datar2004locality}, LSH uses random projections of high-dimensional data to hash samples into different buckets. The LSH algorithm forms the foundation of many ANN search methods, which provide theoretical guarantees and have been extensively studied since its conception \cite{andoni2006near,kulis2009kernelized,andoni2014beyond,andoni2015optimal}. 

Here, we are interested in nearest neighbor retrieval for sets. More precisely, given a training {\it set} of sets as training data (think of {\it set} of point clouds), and a test set (a point cloud representation of an object) we would like to retrieve the ``nearest'' sets in our training {\it set} in an efficient manner. To that end, we extend LSH to enable its application to set retrieval. While there has been a few recent work \cite{nagarkar2018pslsh,kaplan2020locality} on the topic of LSH for set queries, our proposed approach significantly differs from these work. In contrast to \cite{nagarkar2018pslsh,kaplan2020locality}, we provide a Euclidean embedding for sets, which allows for a direct utilization of the LSH algorithm and provides collision probabilities as a function of the set metrics. 

{\bf Wasserstein-based learning:} Wasserstein distances are rooted in the optimal transportation problem \cite{villani2008optimal,kolouri2017optimal,peyre2018computational}, and they provide a robust mathematical framework for comparing probability distributions that respect the underlying geometry of the space.  Wasserstein distances have recently received abundant interest from the Machine Learning and Computer Vision communities. These distances and their variations (e.g., Sliced-Wasserstein distances \cite{rabin2012wasserstein,bonnotte2013unidimensional} and subspace robust Wasserstein distances \cite{paty2019subspace}) have been extensively studied in the context of deep generative modeling \cite{arjovsky2017wasserstein,gulrajani2017improved,tolstikhin2018wasserstein,kolouri2018sliced,liutkus2019sliced}, domain adaptation \cite{courty2016optimal,damodaran2018deepjdot,balaji2019normalized,lee2019sliced}, transfer learning \cite{alvarez2020geometric}, adversarial attacks \cite{wong2019wasserstein,wu2020stronger}, and adversarial robustness \cite{levine2020wasserstein,sinha2018certifying}.

More recently, Wasserstein distances and optimal transport have been used in the context of comparing set-structured data. The main idea behind these recent approaches is to treat sets (with possibly variable cardinalities) as empirical distributions and use transport-based distances for comparing/modeling these distributions. For instance, Togninalli et al. \cite{togninalli2019wasserstein} propose to compare node embeddings of two graphs (treated as sets) via the Wasserstein distance. Later, Mialon et al. \cite{mialon2021a} and Kolouri et al. \cite{kolouri2021wasserstein} concurrently propose Wasserstein embedding frameworks for extracting fixed-dimensional representations from set-structured data. Here, we further extend this direction and propose Sliced-Wasserstein Embedding as a computationally efficient approach that allows us to perform data-independent non-parametric learning from set-structured data.

\section{Preliminaries}
\label{sec:prelim}

We denote an input set with $N_i$ elements living in $\mathbb{R}^d$ by $X_i=\{x_n^i\in\mathbb{R}^d\}_{n=0}^{N_i-1}$. We view sets as empirical probability measures, $\mu_i$, defined in $\mathbb{X}\subseteq\mathbb{R}^d$ with probability density $d\mu_i(x)=p_i(x)dx$, where $p_i(x)=\frac{1}{N_i}\sum_{n=1}^{N_i}\delta(x-x_n^i)$, and $\delta(\cdot)$ is the Dirac delta function. The main idea is then to define the distance between two sets, $X_i$ and $X_j$, as a probability metric between their empirical distributions.

\subsection{Sliced-Wasserstein distances}
\label{subsec:sw}
We use the Sliced-Wasserstein (SW) probability metric as a distance measure between the sets (viewed as empirical probability distributions). Later we will see that this choice allows us to effectively embed sets into a vector space such that the Euclidean distance between embedded sets is equal to the SW-distance between their corresponding empirical distributions. But first, let us briefly define the Wasserstein and Sliced-Wasserstein distances. 

Let $\mu_i$ and $\mu_j$ be one-dimensional probability measures defined on $\mathbb{R}$. Then the p-Wasserstein distance between these measures can be written as:
\begin{equation}
    \mathcal{W}_p(\mu_i,\mu_j)=\big(\int_{0}^{1} (F^{-1}_{\mu_i}(\tau)-F^{-1}_{\mu_j}(\tau))^p d\tau\big)^\frac{1}{p}
\end{equation}
where $F^{-1}_{\mu}$ is the inverse of the cumulative distribution function (c.d.f) $F_\mu$ of $\mu$, i.e., it is the quantile function. The one-dimensional p-Wasserstein distance for empirical distributions with $N$ and $M$ samples can be computed with $\mathcal{O}(Nlog(N)+M log(M))$, which is in stark difference from the generally cubic order for (d$>$1)-dimensional distributions. The one-dimensional case motivates the concept of Sliced-Wasserstein distances \cite{rabin2012wasserstein,kolouri2019generalized}. For the rest of this paper we will consider only the case $p=2$, and for brevity we refer to 2-Wasserstein and 2-Sliced-Wasserstein distances as Wasserstein and Sliced-Wasserstein distances. 

The main idea behind SW distances is to slice d-dimensional distributions into infinite sets of their one-dimensional slices/marginals and then calculate the expected Wasserstein distance between their slices. Let $g_\theta: \mathbb{R}^d \rightarrow \mathbb{R}$ be a parametric function with parameters $\theta\in\Omega_\theta\subseteq\mathbb{R}^{d_\theta}$, satisfying the regularity conditions in both inputs and parameters as presented in \cite{kolouri2019generalized}. A common choice is $g_\theta(x)=\theta^Tx$ where $\theta\in\mathbb{S}^{d-1}$ is a unit vector in $\mathbb{R}^d$, and $\mathbb{S}^{d-1}$ denotes the unit $d$-dimensional hypersphere. The slice of a probability measure, $\mu$, with respect to $g_\theta$ is the one-dimensional probability measure $g_{\theta\#}\mu$, with the density,
\begin{align}\label{eq:sliced_measure}
    p^\theta_i(t):= \int_{\mathbb{X}} \delta(t-g_\theta(x))d\mu_i(x)~~~\forall t\in\mathbb{R}.
\end{align}
The generalized Sliced-Wasserstein distance is defined as
{\small
\begin{align}
    \mathcal{GSW}_2(\mu_i,\mu_j)=\left(\int_{\Omega_\theta} \mathcal{W}^2_2(g_{\theta\#}\mu_i,g_{\theta\#}\mu_j)d\sigma(\theta) \right)^{\frac{1}{2}},
    \label{eq:GSW}
\end{align}}
where $\sigma(\theta)$ is the uniform measure on $\Omega_\theta$, and once again for $g_\theta(x)=\theta^T x$ and $\Omega_\theta=\mathbb{S}^{d-1}$, the generalized Sliced-Wasserstein distance is simply the Sliced-Wasserstein distance. Equation \eqref{eq:GSW} is the expected value of the Wasserstein distances between uniformly distributed slices (i.e., on $\Omega_\theta$) of distributions $\mu_i$ and $\mu_j$.

\subsection{Locality Sensitive Hashing}
Locality Sensitive Hashing (LSH) \cite{indyk1998approximate,datar2004locality} strives for hashing near points in the high-dimensional input-space into the same hash bucket with a higher probability than distant ones, effectively solving the $(R,c)$-Near
Neighbor problem. More precisely, let $H:=\{h:\mathbb{R}^d\rightarrow U\}$ denote a LSH function family with $U$ denoting the hash values. The LSH function family $H$, is called $(R,c,P_1,P_2)$-sensitive if for any two points $u,v\in \mathbb{R}^d$ and $\forall h\in H$ it satisfies the following conditions:
\begin{itemize}
    \item If $\|u-v\|_2\leq R$, then $\operatorname{Pr}[h(u)=h(v)]\geq P_1$, and
    \item If $\|u-v\|_2> cR$, then $\operatorname{Pr}[h(u)=h(v)]\leq P_2$
\end{itemize}
Where for a proper LSH $c>1$ and $P_1>P_2$. The original LSH for Euclidean distance uses,
\begin{align*}
    h_{a,b}(u)=\lfloor\frac{a^Tu+b}{\omega}\rfloor
\end{align*}
where $\lfloor\cdot\rfloor$ is the floor function, $a\in\mathbb{R}^d$ is a random vector generated from a $p$-stable distribution (e.g., Normal or Cauchy distributions) \cite{datar2004locality}, and $b$ is a real number chosen uniformly from $[0,\omega)$ such that $\omega$ is the width of the hash bucket. Let $r=\|u-v\|_p$ and $f_p(t)$ denote the probability density function of the absolute value of the p-stable distribution, then this family of hash functions leads to the following probability of collision:
\begin{equation}
    \operatorname{Pr}[h_{a,b}(u)=h_{a,b}(v)]=\int_0^\omega \frac{1}{r}f_p(\frac{t}{r})(1-\frac{t}{\omega})dt
\end{equation}
To amplify the gap between $P_1$ and $P_2$ one can concatenate several such hash functions. Let $\mathcal{G}:=\{g:\mathbb{R}^d\rightarrow U^k\}$ denote the concatenation of $k$ randomly selected hash functions, i.e.,  $g(u)=(h_1(u),...,h_k(u))$. In practice, multiple such hash functions $g_1,\ldots,g_T$ are often used. For points, $u$ and $v$, within $R$-distance from one another, the probability that they collide at least in one of $g_j$s is: $1-(1-P_1^k)^T$. 

Another commonly used, and related, family of hash functions consist of random projections and thresholding, i.e., $h_{a,b}(u)=\operatorname{sgn}(a^Tu+b)$ \cite{charikar2002similarity}. Using $k$ such random projections (i.e., binary codes of length $k$) provides the following collision probability: 

$$ \operatorname{Pr}[g(u)=g(v)]=[1-\frac{\cos^{-1}(u^Tv)}{\pi}]^k$$

Since the conception of its idea \cite{indyk1998approximate,datar2004locality}, many variants of LSH have been proposed. However, these approaches are not designed to handle set queries. Here, we extend LSH to set-structured data via Sliced-Wasserstein Embeddings. 

\begin{figure*}[t!]
    \centering
    \includegraphics[width=\linewidth]{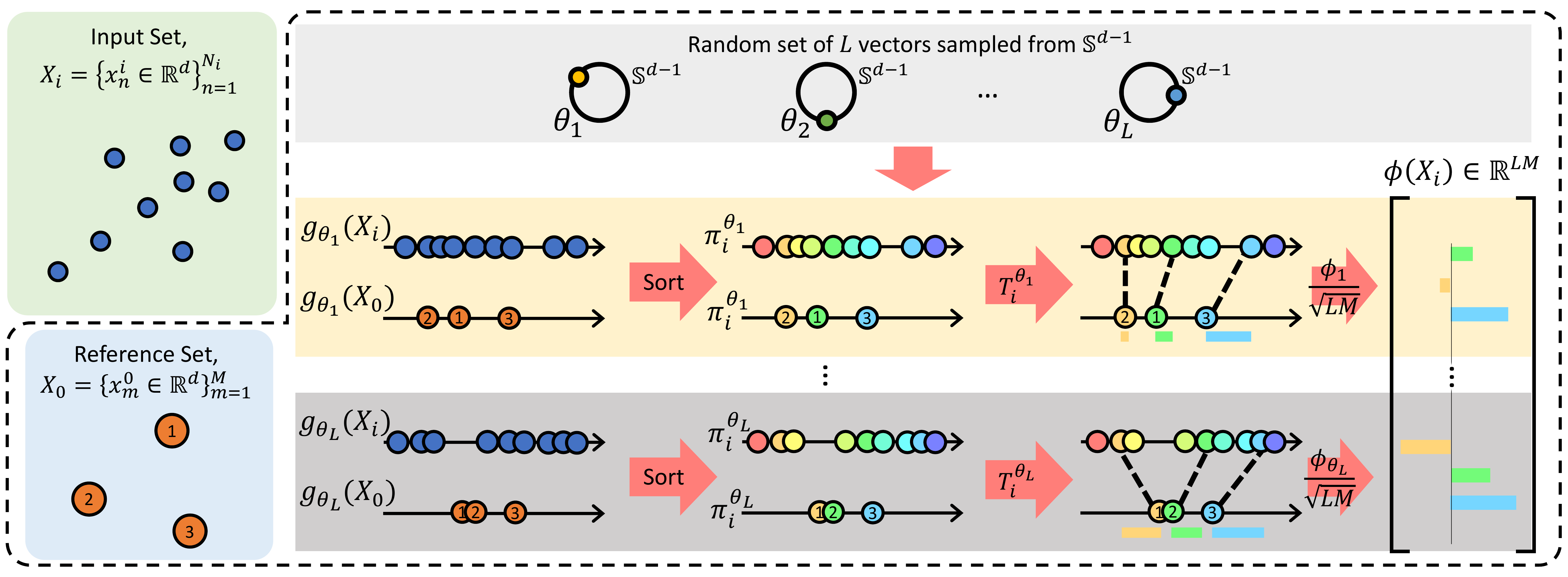}
    \caption{A graphical depiction of Sliced-Wasserstein Embedding (SWE) for a given reference set and a chosen number of slices, $L$. The input and reference sets are sliced via $L$ random projections $\{g_{\theta_l}\}_{l=1}^L$. The projections are then sorted and the Monge couplings between input and reference sets' slices are calculated following Eq.\ \eqref{eq:mongeCoupling}. Finally, the embedding is obtained by weighting and concatenating $\phi_{\theta_l}$s. The Euclidean distance between two embedded sets is equal to their corresponding $\widehat{\mathcal{GSW}}_{2,L}$ distance/dissimilarity measure, i.e., $\|\phi(X_i)-\phi(X_j)\|_2=\widehat{\mathcal{GSW}}_{2,L}(X_i,X_j)\approx \mathcal{GSW}_2(X_i,X_j)$. }
    \label{fig:swe}
\end{figure*}

\section{Problem Formulation and Method}
\label{sec:method}

\subsection{Sliced-Wasserstein Embedding}

The idea of Sliced-Wasserstein Embeddings (SWE) is rooted in Linear Optimal Transport \cite{wang2013linear,kolouri2017optimal,moosmuller2020linear} and was first introduced in the context of pattern recognition from 2D probability density functions (e.g., images) \cite{kolouri2016radon} and more recently in \cite{shifat2021radon}. Our work extends the framework to d-dimensional distributions with the specific application of set retrieval in mind. Consider a set of probability measures $\{\mu_i\}_{i=1}^I$ with densities $\{p_i\}_{i=1}^I$, and recall that we use $\mu_i$ to represent the i'th set $X_i=\{x_n^i\}_{n=0}^{N_i-1}$. At a high level, SWE can be thought as a set-2-vector operator, $\phi$, such that:
\begin{align}
\|\phi(\mu_i)-\phi(\mu_j)\|_2=\mathcal{GSW}_2(\mu_i,\mu_j).
\label{eq:euclidean}
\end{align}

 For convenience, we use $\mu_i^\theta:= g_{\theta\#} \mu_i$ to denote the slice of measure $\mu_i$ with respect to $g_{\theta}$. Also, let $\mu_0$ denote a reference measure, with $\mu_0^\theta$ being its corresponding slice. The optimal transport coupling (i.e., Monge coupling) between $\mu_i^\theta$ and $\mu_0^\theta$ can be written as
\begin{align}
    T^\theta_i = F_{\mu_i^\theta}^{-1}\circ F_{\mu_0^\theta},
    \label{eq:monge}
\end{align}
and recall that $F_{\mu_i^\theta}^{-1}$ is the quantile function of $\mu_i^\theta$. Now, letting $id$ denote the identity function, we define the so-called cumulative distribution transform (CDT) \cite{park2018cumulative} of $\mu_i^\theta$ as
\begin{align}\label{eq:cdt}
\phi_\theta(\mu_i)\coloneqq (T_i^\theta - id).
\end{align}
For a fixed $\theta$, we can show that $\phi^\theta(\mu_i)$ satisfies (see supplementary material):
\begin{enumerate}
    \item [\bf C1.]  The weighted $\ell_2$-norm of the embedded slice, $\phi_\theta(\mu_i)$, satisfies:
    \begin{align*}
        \|\phi_\theta(\mu_i)\|_{\mu_0^\theta,2}&=\left(\int_{\mathbb{R}}\|\phi_\theta(\mu_i(t))\|_2^2 d\mu_0^\theta(t)\right)^{\frac{1}{2}} \\&=\mathcal{W}_2(\mu_i^\theta,\mu_0^\theta),
    \end{align*}
    As a corollary we have $\|\phi_\theta(\mu_0)\|_{\mu_0^\theta,2}=0$.
    \item [\bf C2.] The weighted $\ell_2$ distance between two embedded slices satisfies:
    \begin{align}
        \|\phi_\theta(\mu_i)-\phi_\theta(\mu_j)\|_{\mu_0^\theta,2}=\mathcal{W}_2(\mu_i^\theta,\mu_j^\theta).
        \label{eq:weightedEuclidean}
    \end{align}
\end{enumerate}
It follows from {\bf C1} and {\bf C2} that:
{\small
\begin{align}
\label{eq:embedding} 
\mathcal{GSW}_2(\mu_i,\mu_j)&=\left(\int_{\Omega_\theta}\|\phi_\theta(\mu_i)-\phi_\theta(\mu_j)\|_{\mu_0^\theta, 2}^2 d\sigma(\theta)\right)^{\frac{1}{2}}
\end{align}}
For probability measure $\mu_i$, we then define the mapping to the embedding space via, $$\phi(\mu_i)\coloneqq\{\phi_\theta(\mu_i)~|~\theta \in \Omega_\theta \}.$$
Finally, we can re-weight the embedding (according to $d\mu^\theta_0$) such that the weighted $\ell_2$ in \ref{eq:weightedEuclidean} becomes the $\ell_2$ distance as in \ref{eq:euclidean}. Next, we describe this reweighting and other implementation considerations in more details. 
\subsection{Monte Carlo Integral Approximation}

The GSW distance in Eq.\ \eqref{eq:GSW} relies on integration on $\Omega_\theta$ (e.g., $\mathbb{S}^{d-1}$ for linear slices), which cannot be directly calculated. Following the common practice in the literature \cite{rabin2012wasserstein,bonneel2015sliced,kolouri2016sliced,kolouri2018sliced,csimcsekli2018sliced}, 
here, we approximate the integration on $\theta$ via a Monte-Carlo (MC) sampling of $\Omega_\theta$. Let $\Theta_L=\{\theta_l\sim \sigma(\theta)\}_{l=1}^L$ denote a set of $L$ parameters sampled independently and uniformly from $\Omega_\theta$. We assume an empirical reference measure,  $\mu_0=\frac{1}{M}\sum_{m=1}^M \delta(x-x_m^0)$ with $M$ samples. The MC approximation is written as:
\begin{align}
    \widehat{\mathcal{GSW}}^2_{2,L}(\mu_i,\mu_j)=\frac{1}{LM}\sum_{l=1}^{L} \| \phi_{\theta_l}(\mu_i)-\phi_{\theta_l}(\mu_j)\|^2_2. 
\end{align}
Finally, our SWE embedding is calculated via: 
\begin{align}
    \phi(\mu_i)=[\frac{\phi_{\theta_1}(\mu_i)}{\sqrt{LM}};...;\frac{\phi_{\theta_L}(\mu_i)}{\sqrt{LM}}]\in\mathbb{R}^{LM\times 1},
    \label{eq:fembedd}
\end{align}
 which satisfies:
$$\|\phi(\mu_i)-\phi(\mu_j)\|_2= \widehat{\mathcal{GSW}}_{2,L}(\mu_i,\mu_j) \approx \mathcal{GSW}_2(\mu_i,\mu_j).$$
As for the approximation error, we rely on Theorem 6 in \cite{nadjahi2020statistical}, which uses Hölder’s inequality and the results on the moments of the Monte Carlo estimation error to obtain:
{\scriptsize
\begin{align}
\mathbb{E}\big[|\widehat{\mathcal{GSW}}^2_{2,L}(\mu_i,\mu_j)-\mathcal{GSW}^2_{2}(\mu,\nu)|\big]\leq \sqrt{\frac{\mathrm{var}(\mathcal{W}_2^2(\mu^\theta_i,\mu^\theta_j))}{L}}
\end{align}}
The upper bound indicates that the approximation error decreases with $\sqrt{L}$. The numerator, however, is implicitly dependent on the dimensionality of input space. Meaning that a larger number of slices, $L$, is needed for higher dimensions.

\subsection{SWE Algorithm}
Here we review the algorithmic steps to obtain SWE. We consider $X_i=\{x^i_n\}_{n=0}^{N_i-1}$ as the input set with $N_i$ elements, and $X_0=\{x^0_m\}_{m=0}^{M-1}$ denote the  reference set of $M$ samples where in general $M\neq N_i$. 
For a fixed slicer $g_\theta$ we calculate $\{g_\theta(x^i_n)\}_{n=0}^{N_i-1}$ and $\{g_\theta(x^0_m)\}_{m=0}^{M-1}$ and sort them increasingly. Let $\pi_i$ and $\pi_0$ denote the permutation indices (obtained from argsort). Also, let $\pi_0^{-1}$ denote the ordering that permutes the sorted set back to the original ordering. Then we numerically calculate the Monge coupling $T^\theta_i$ via:
\begin{equation}
T^\theta_i[m]=F_{\mu_i^{\theta}}^{-1}\left(\frac{\pi_0^{-1}[m]+1}{M}\right)    
\label{eq:mongeCoupling}
\end{equation}
where $F_{\mu_0^{\theta}}(x_m^0)=\frac{\pi^{-1}_0[m]+1}{M}$, assuming that the indices start from 0. Here $F_{\mu_i^{\theta}}^{-1}$ is calculated via interpolation. In our experiments we used linear interpolation similar to \cite{liutkus2019sliced}. Note that the dimensionality of the Monge coupling is only a function of the reference cardinality, i.e., $T_i^\theta\in\mathbb{R}^M$. Consequently, we write:
\begin{equation}
    \phi_\theta(X_i)[m]=(T_i^\theta[m]-g_{\theta}(x^0_m))
    \label{eq:embedding}
\end{equation}
and repeat this process for $\theta \in \{\theta_l\sim\sigma(\theta)\}_{l=1}^L$, while we emphasize that this process can be parallelized.
The final embedding is achieved via weighting and concatenating $\phi_{\theta_l}$s as in Eq. \eqref{eq:fembedd}, where the coefficient $\frac{1}{\sqrt{LM}}$ allows us to simplify the weighted Euclidean distance, $\|\cdot\|_{\mu_0,2}$, to Euclidean distance, $\|\cdot\|_{2}$. Algorithm \ref{alg:swe} summarizes the embedding process, and Figure \ref{fig:swe} provides a graphical depiction of the process. Lastly, the SWE's computational complexity for a set with cardinality $|X|=N$ is  $\mathcal{O}(LN(d+logN))$, where we assumed the cardinality of the reference set is of the same order as $N$. Note that $\mathcal{O}(LNd)$ is the cost of slicing and $\mathcal{O}(LNlog(N))$ is the sorting and interpolation cost to calculate Eq. \eqref{eq:mongeCoupling}.

\begin{algorithm}[t!]
\caption{Sliced-Wasserstein Embedding}
\begin{algorithmic}
\Procedure{SWE}{$X_i=\{x_n^i\}_{n=0}^{N_i-1}$, $X_0=\{x_m^0\}_{m=0}^{M-1}$, $L$}
    \State Generate a set of $L$ samples $\Theta_L=\{\theta_l\sim \mathcal{U}_{\Omega_\theta}\}_{l=1}^L$
    \State Calculate $g_{\Theta_L}(X_0):=\{g_{\theta_l}(x_m^0)\}_{m,l}\in\mathbb{R}^{M\times L}$
    \State Calculate $\pi_0=\mbox{argsort}(g_{\Theta_L}(X_0))$ and $\pi_0^{-1}$ (on $m$-axis)
    \State Calculate $g_{\Theta_L}(X_i):=\{g_{\theta_l}(x_n^i)\}_{n,l}\in\mathbb{R}^{N_i\times L}$
    \State Calculate $\pi_i=\mbox{argsort}(g_{\Theta_L}(X_i))$ (on $n$-axis)
    \For{$l=1$ to $L$}
    \State Calculate the Monge coupling $T_i^{\theta_l}\in \mathbb{R}^{M}$ (Eq. \eqref{eq:mongeCoupling})
    \EndFor
    \State Calculate the embedding $\phi(X_i)\in \mathbb{R}^{M\times L}$ (Eq. \eqref{eq:fembedd})
    \State \Return $\phi(X_i)$
   \EndProcedure
    \end{algorithmic}
\label{alg:swe}
\end{algorithm}

\subsection{SLOSH}
Our proposed Set LOcality Sensitive Hashing (SLOSH) leverages the Sliced-Wasserstein Embedding (SWE) to embed training sets, $\mathcal{X}_{train}=\{X_i | X_i=\{x^i_{n}\in \mathbb{R}^d \}_{n=0}^{N_i-1}\}$, into a vector space where we can use Locality Sensitive Hashing (LSH) to perform ANN search. We treat each input set $X_i$ as a probability measure $\mu_i(x)=\frac{1}{N_i}\sum_{n=1}^{N_i} \delta(x-x_n^i)$. For a reference set $X_0$ with cardinality $|X_0|=M$ and $L$ slices, we embed the input set $X_i$ into a $(\mathbb{R}^{LM})$-dimensional vector space using Algorithm \ref{alg:swe}. With abuse of notation we use $\phi(\mu_i)$ and $\phi(X_i)$ interchangeably. 

Given SWE, a family of $(R,c,P_1,P_2)$-sensitive LSH functions, $H$, will induce the following conditions,
\begin{itemize}
    \item If $\widehat{\mathcal{GSW}}_{2,L}(X_i,X_j)\leq R$, then $Pr[h(\phi(X_i))=h(\phi(X_j))]\geq P_1$, and
    \item If $\widehat{\mathcal{GSW}}_{2,L}(X_i,X_j)> cR$, then $Pr[h(\phi(X_i))=h(\phi(X_j))]\leq P_2$
\end{itemize}
As mentioned in Section \ref{sec:related}, for amplifying the gap between $P_1$ and $P_2$, one uses $g(X_i)=[h_1(\phi(X_i)),...,h_k(\phi(X_i))]$, which results in a code length, $k$, for each input set, $X_i$.  Finally, if $\widehat{\mathcal{GSW}}_{2,L}(X_i,X_j)\leq R$, by using $T$ such codes, $g_t$ for $t\in\{1,...,T\}$, of length $k$, we can ensure collision at least in one of $g_t$s  with probability $1-(1-P_1^k)^T$.

\begin{table*}[t!]
\caption{Retrieval results of baselines and our approach on set retrieval on the three data sets.}
\resizebox{\textwidth}{!}{%
\begin{tabular}{l c c c c c c c c c}
\hline
& \multicolumn{3}{c}{Point MNIST 2D} & \multicolumn{3}{c}{ModelNet 40} & \multicolumn{3}{c}{Oxford 5k}\\
\hline
&\multicolumn{3}{c}{Precision@k/Accuracy} & \multicolumn{3}{c}{Precision@k/Accuracy}& 
\multicolumn{3}{c}{Precision@k/Accuracy}
\\
& k=4 & k=8 & k=16 & k=4 & k=8 & k=16 & k=4 & k=8 & k=16 \\ 
\hline
GeM-1 (GAP) & 
0.10/0.11  & 0.10/0.10 & 0.10/0.10 & 0.16/0.19 & 0.16/0.22 & 0.16/0.25 & 0.29/0.35 & 0.25/0.31 & 0.22/0.29 \\ 

GeM-2 & 
0.29/0.32 & 0.28/0.35 & 0.29/0.37 & 0.29/0.34 & 0.26/0.34 & 0.23/0.33 & 0.38/0.53 & 0.31/0.40 & 0.27/0.38 \\ 

GeM-4 & 
0.39/0.45 & 0.39/0.47 & 0.38/0.49 & 0.31/0.37 & 0.28/0.38 & 0.25/0.37 & 0.35/0.41 & 0.33/0.51 & 0.29/0.42 \\ 

Cov & 
0.25/0.27 & 0.25/0.28 & 0.25/0.28 & 0.37/0.42 & 0.35/0.44 & 0.32/0.44 & 0.35/0.55 & 0.30/0.37 & 0.26/0.33\\ 

FSPool &
\textbf{0.75/0.80} & \textbf{0.74/0.81} & \textbf{0.73/0.81} & 
\textbf{0.50/0.57} & \textbf{0.47/0.58} & \textbf{0.42/0.56} & 
\textbf{0.47/0.53} & \textbf{0.39/0.6} & \textbf{0.32/0.49} \\ 

\hline

SLOSH ($L<d$)
& 0.52/0.59 & 0.51/0.61 & 0.49/0.61 
& 0.22/0.25 & 0.20/0.27 & 0.19/0.27 
& 0.33/0.42 & 0.30/0.49 & 0.24/0.36  \\ 

SLOSH ($L=d$)
& \textbf{0.67/0.73} & \textbf{0.64/0.74} & \textbf{0.62/0.73} & \textbf{0.39/0.44} & \textbf{0.36/0.45} & \textbf{0.33/0.44} & 
\textbf{0.43/0.58} & \textbf{0.39/0.60} & \textbf{0.34/0.55} \\ 

SLOSH ($L>d$)
& \textbf{0.90/0.92} & \textbf{0.88/0.92} & \textbf{0.87/0.91} & \textbf{0.55/0.61} & \textbf{0.51/0.60} & \textbf{0.46/0.57} & \textbf{0.53/0.69} & \textbf{0.45/0.65} & \textbf{0.38/0.64} \\ 
\hline
\end{tabular}
}
\label{tab:main}
\end{table*}

\section{Experiments}
\label{sec:experiments}

We evaluated our set locality sensitive hashing via Sliced-Wasserstein Embedding algorithm against using Generalized Mean (GeM) pooling \cite{radenovic2018fine}, Global Covariance (Cov) pooling  \cite{wang2020deep}, and Featurewise Sort Pooling (FSPool)  \cite{zhang2019fspool} on various set-structured datasets. We note that while FSPool was proposed as a data-dependent embedding, here we devise its data-independent variation for fair comparison. Interstingly, FSPool can be thought as a special case of our SWE embedding where $L=d$ and $\Theta_L$ is chosen as the identity matrix. To evaluate these methods, we tested all approaches on point cloud MNIST dataset (2D) \cite{lecun1998gradient,pointCloudMNIST}, ModelNet40 dataset (3D) \cite{wu20153d}, and the Oxford Buildings dataset \cite{philbin2007object} (i.e., Oxford 5k). 

\subsection{Baselines}
Let ${X}_i=\{x^i_n\in\mathbb{R}^d\}_{n=0}^{N_i-1}$ be the input set with $N_i$ elements. We denote $[X_i]_{k}=\{[x^i_n]_{k}\}_{n=0}^{N_i-1}$ as the set of all elements along the $k$'th dimension, $k \in \{1, 2, ...d\}$. Below we provide a quick overview of the baseline approaches, which provide different set-2-vector mechanisms. 

{\bf Generalized-Mean Pooling (GeM)} \cite{radenovic2018fine} was originally proposed as a generalization of global mean and global max pooling on Convolutional Neural Network (CNN) features to boost image retrieval performance. Given the input ${X}_i$, GeM calculates the (p-th)-moment of each feature, $f^{(p)}\in\mathbb{R}^d$, as: 

\begin{equation}
    [f^{(p)}]_{k} = \left(\frac{1}{N_i}\sum_{n=1}^{N_i} ([x^i_n]_{k})^{p}\right)^{\frac{1}{p}}
    \label{eq:GeM}
\end{equation}
When pooling parameter $p=1$, we end up with average pooling. While as $p \to \infty$, we get max pooling. In practice, we found that a concatenation of higher-order GeM features, i.e., $\phi_{GeM}(X_i)=[f^{(1)};...;f^{(p)}]\in\mathbb{R}^{pd}$, leads to the best performance, where $p$ is GeM's hyper-parameter.

{\bf Covariance Pooling} \cite{acharya2018covariance,wang2020deep} presents another way to capture second-order statistics and provide more informative representations. It was shown that this mechanism can be applied to CNN features as an alternative to global mean/max pooling to generate state-of-the-art results on facial expression recognition tasks\cite{acharya2018covariance}. Given input set ${X}_i$, the unbiased covariance matrix is computed by:
\begin{align}
    C = \frac{1}{N_i - 1}\sum_{n=1}^{N_i} (x^i_n - \overline{\mu}_i)(x^i_n - \overline{\mu}_i)^T,
\end{align}
where $\overline{\mu}_i = \frac{1}{N_i} \sum_{n=1}^{N_i} x^i_n$. The output matrix can be further regularized by adding a multiple of the trace to diagonal entries of the covariance matrix to ensure symmetric positive definiteness (SPD), $C_{\lambda} = C + \lambda trace(C)\textbf{I}$ where $\lambda$ is a regularization hyper-parameter and \textbf{I} is the identity matrix. Covariance pooling then uses $\phi_{Cov}(X_i)=\text{flatten}(C_\lambda)$.

{\bf Featurewise Sort Pooling (FSPool)} \cite{zhang2019fspool} is a powerful technique for learning representations from set-structured data. In short, this approach is based on sorting features along all elements of a set, $[X_i]_k$:
\begin{equation}
    f = [Sorted([X_i]_{1}), ..., Sorted([X_i]_{d}))]\in \mathbb{R}^{N_i\times d}
\end{equation}
The fixed-dimensional representation is then obtained via an interpolation along the $N_i$ dimension of $f$. More precisely, a continuous linear operator $W$ is learned and the inner product between this continuous operator and $f$ is evaluated at $M$ fixed points (i.e., leading to weighted summation over $d$), resulting in a $M$-dimensional embedding.
 
Given that we are interested in a data-independent set representation, we cannot rely on learning the continuous linear operator $W$. Instead, we perform interpolation along the $N_i$ axis on $M$ points and drop the inner product all together to obtain a $(M\times d)$-dimensional data-independent set representation. The mentioned variation of FSPool is similar to the Sliced-Wasserstein Embedding when $L=d$ and $\Theta_L=I_{d\times d}$, i.e., axis-aligned projections.   




\subsection{Datasets}
\label{subsec:exp1}
{\bf Point Cloud MNIST 2D} \cite{lecun1998gradient,pointCloudMNIST} consists of 60,000 training samples and 10,000 testing samples. Each sample is a 2-dimensional point cloud derived from an image in the original MNIST dataset. The sets have various cardinalities in the range of $|X_i|\in[34,351]$. 

{\bf ModelNet40} \cite{wu20153d} contains 3-dimensional point clouds converted from 12,311 CAD models in 40 common object categories. We used the official split with 9,843 samples for training and 2,468 samples for testing. We sample $N_i$ points uniformly and randomly from the mesh faces of each object, where $N_i=\lfloor n_i \rfloor,  n_i\sim\mathcal{N}(512,64)$. To avoid any orientation bias, we randomly rotate the point clouds by $\rho\in[0,45]$ degrees around the x-axis. Finally, we normalize each sample to fit within the unit cube to avoid any scale bias and to enforce the methods to rely on shapes. 

{\bf The Oxford Buildings Dataset} \cite{philbin2007object} has 5,062 images containing eleven different Oxford landmarks. Each landmark has five corresponding queries leading to 55 queries over which an object retrieval system can be evaluated. In our experiments, we use a pretrained VGG16 \cite{simonyan2014very} on ImageNet1k \cite{ILSVRC15} as a feature extractor, and use the features in the last convolutional layer as a set representation for an input image. We resize the images without cropping, which leads to varied input size images, and therefore gives set representations with varied cardinalities. We further perform a dimensionality reduction on these features to obtain sets of features in an $(d=8)$-dimensional space. 

\begin{figure}[t!]
    \centering
    \vspace{-.3in}
    \includegraphics[width=\columnwidth]{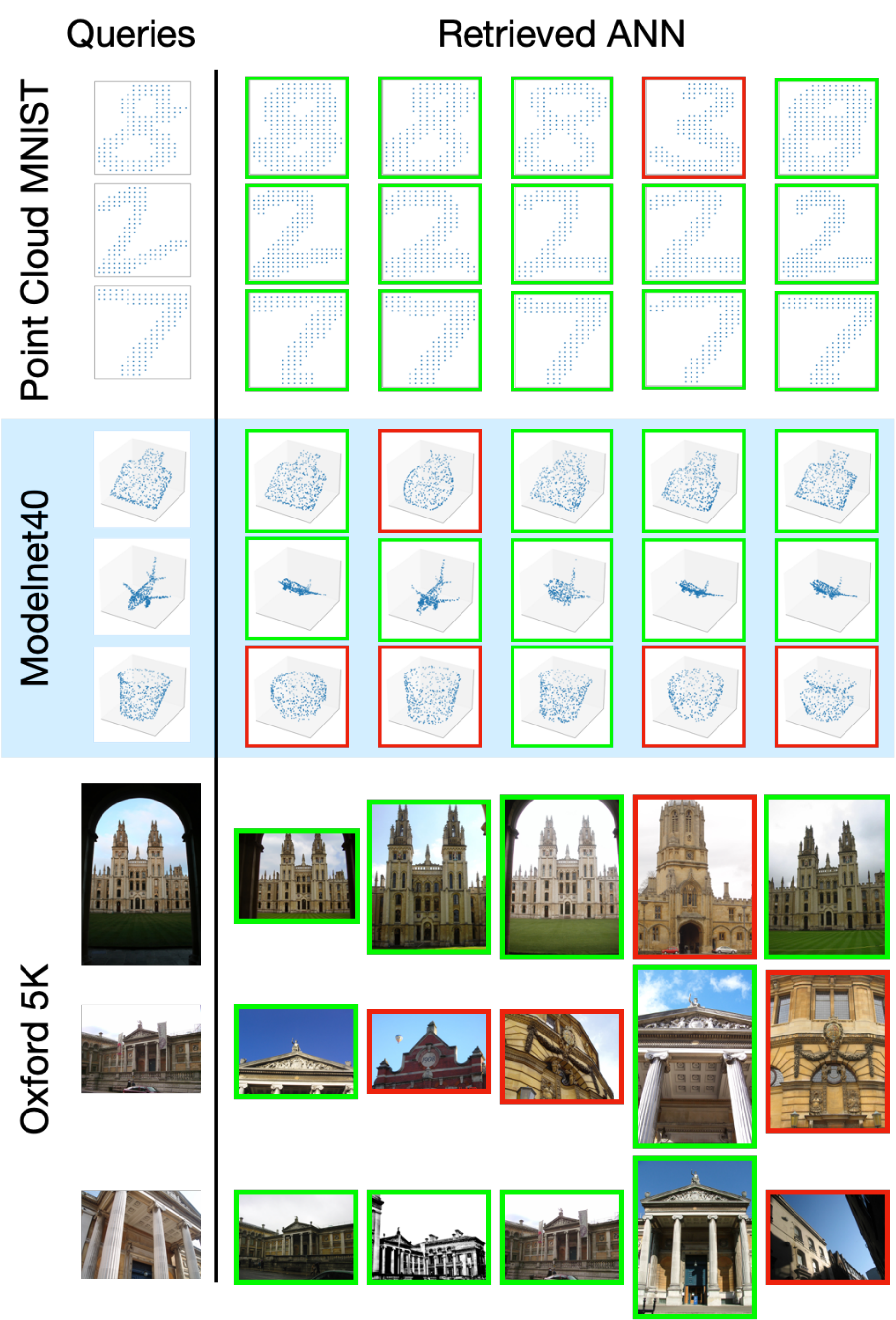}
    \caption{Example sample retrievals.}
    \label{fig:samples}
    \vspace{-.3in}
\end{figure}

\begin{figure}[t!]
    \centering
    \includegraphics[width=\linewidth]{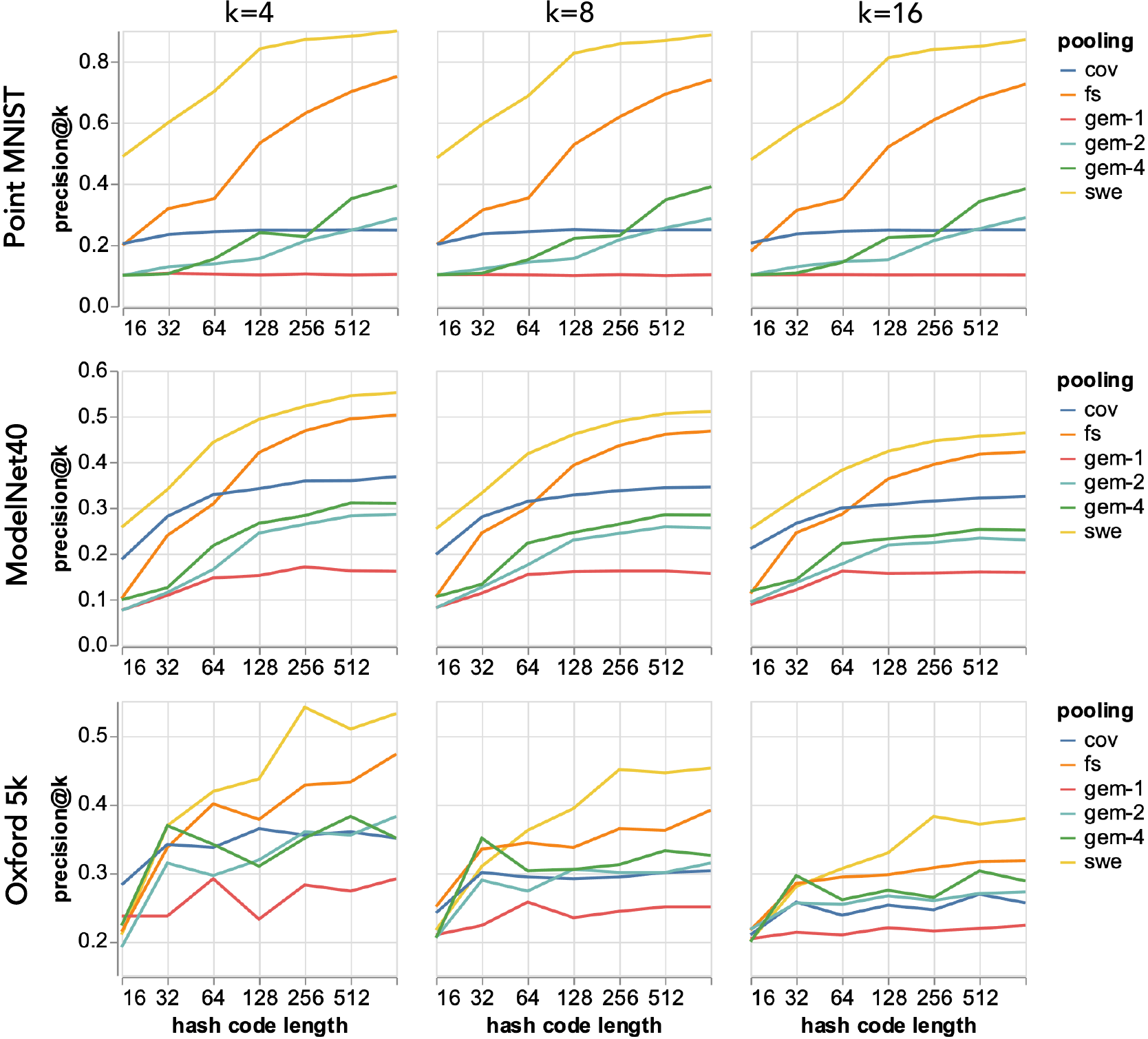}
    \caption{Sensitivity to hash code length. }
    \label{fig:code_length}
    \vspace{-.15in}
\end{figure}

\subsection{Results}

For all datasets, we first calculate the set-2-vector embeddings for all baselines and SWE. Then, we apply Locality-Sensitive Hashing (LSH) to the embedded sets and report Precision@k and accuracy (based on majority voting) for all the approaches on the test sets. We use the FAISS library \cite{JDH17}, developed by Facebook AI Research, for LSH. For all methods, we use a hash code length of $1024$, and we report our results  for $k=4, 8, \text{and}~16$. For SLOSH, we consider three different settings for the number of slices, namely, $L<d$, $L=d$, and $L>d$. We repeat the experiments five times per method and report the mean Precision@k and accuracy in Table \ref{tab:main}. In all experiments, the best hyper-parameters are selected for each approach based on cross validation. We see that SLOSH provides a consistent lead on all datasets, especially, when $L>d$. Figure \ref{fig:samples} provide set retrieval examples on the three data sets. Next, we provide a sensitivity analysis of our approach.

{\bf Sensitivity to code length.} For all datasets, we study the sensitivity of the different embeddings to the hashing code-length. We vary the code length from $16$ to $1024$, and report the average of Precision@k over five runs. Figure \ref{fig:code_length} shows the outcome of this study. It can be seen that methods that encode higher statistical orders of the underlying distribution gain performance as a function of the hash code length. In addition, we see a strong performance gap between SWE and other set-2-vector approaches, which points at the more descriptive nature of our proposed method.  

\begin{figure}[t!]
    \centering
    \includegraphics[width=\linewidth]{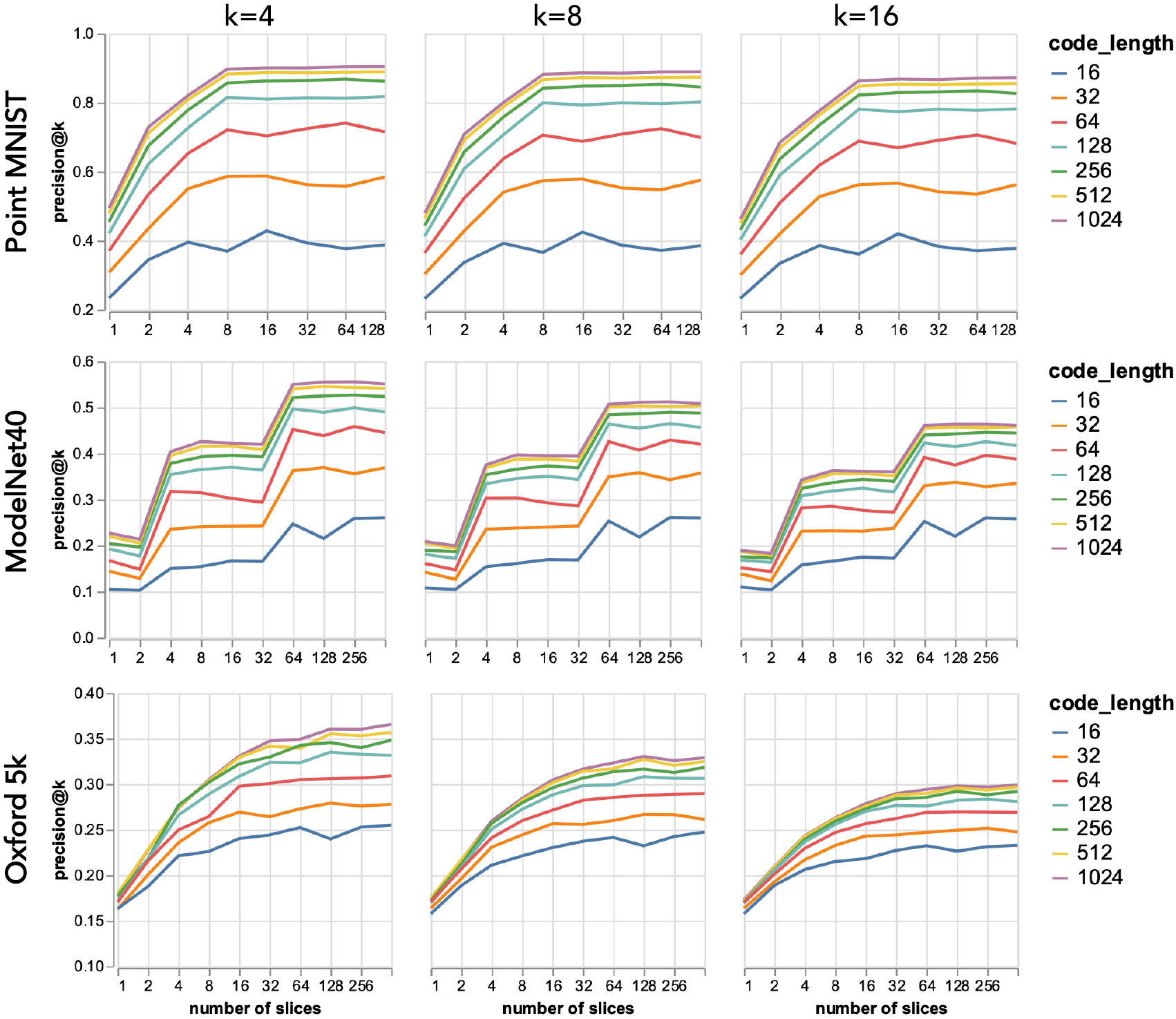}
    \caption{Sensitivity to number of slices. }
    \label{fig:num_slices}
\end{figure}

{\bf Sensitivity to the number of slices.} Next, we study the sensitivity of SLOSH to the choice of number of slices, $L$, for the three datasets. We measure the average Precision@k over five runs for various number of slices and for different code-lengths and report our results in Figure \ref{fig:num_slices}. As expected we see a performance increase with respect to increasing $L$. 

{\bf Sensitivity to the reference.} Finally, we perform a study on the sensitivity to the choice of the reference set, $X_0$. We measure the performance of SLOSH on the three datasets and for various code-lengths, when the reference set is: 1) calculated using K-Means on the elements of all sets, 2) a random set from the dataset, 3) sampled from a uniform distribution, and 4) sampled from the normal distribution. We see that SLOSH's performance depends on the reference choice. However, we point out that our formulation implies that as $L\rightarrow \infty$ the embedding becomes independent of the choice of the reference, and the observed sensitivity could be due to using finite $L$.

\begin{figure}
    \centering
    \includegraphics[width=\linewidth]{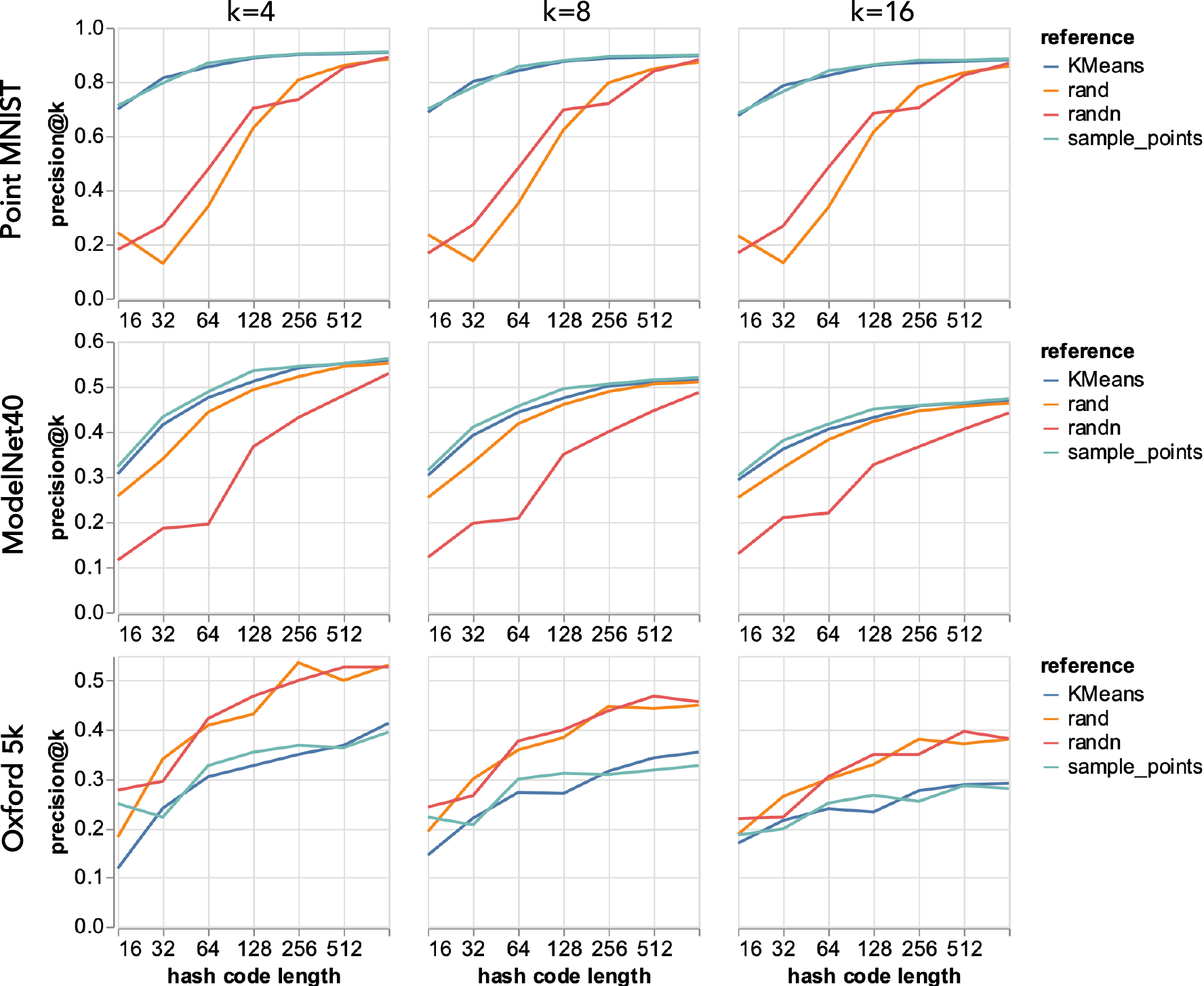}
    \caption{Sensitivity to reference functions. }
    \label{fig:ref_funcs}
    \vspace{-.15in}
\end{figure}

\section{Conclusion}
\label{sec:conclusion}

We described a novel data-independent approach for Approximate Nearest Neighbor (ANN) search on set-structured data, with applications in set retrieval. We treat set elements as samples from an underlying distribution, and embed sets into a vector space in which the Euclidean distance approximates the Sliced-Wasserstein (SW) distance between the input distributions. We show that for a set $X$ with cardinality $|X|=N$, our framework requires $\mathcal{O}(LN(d+Log(N)))$ (sequential processing) or $\mathcal{O}(N(d+log(N)))$ (parallel processing) calculations to obtain the embedding. We then use Locality Sensitive Hashing (LSH) for fast retrieval of nearest sets in our proposed embedding. We demonstrate a significant boost over other data-independent approaches including Generalized Mean (GeM) on three different set retrieval tasks, namely, Point Cloud MNIST, ModelNet40, and the Oxford Buildings datasets. Finally, our proposed method is readily extendable to data-dependent settings by allowing optimization on the slices and the reference set.   
\section{Acknowledgements}

This research was supported by the Defense Advanced Research Projects Agency (DARPA) under Contract No. HR00112190132.

{\small
\bibliographystyle{ieee_fullname}
\bibliography{slosh}

\begin{thebibliography}{10}\itemsep=-1pt

\bibitem{acharya2018covariance}
Dinesh Acharya, Zhiwu Huang, Danda Pani~Paudel, and Luc Van~Gool.
\newblock Covariance pooling for facial expression recognition.
\newblock In {\em Proceedings of the IEEE Conference on Computer Vision and
  Pattern Recognition Workshops}, pages 367--374, 2018.

\bibitem{alvarez2020geometric}
David Alvarez~Melis and Nicolo Fusi.
\newblock Geometric dataset distances via optimal transport.
\newblock {\em Advances in Neural Information Processing Systems}, 33, 2020.

\bibitem{andoni2006near}
Alexandr Andoni and Piotr Indyk.
\newblock Near-optimal hashing algorithms for approximate nearest neighbor in
  high dimensions.
\newblock In {\em 2006 47th annual IEEE symposium on foundations of computer
  science (FOCS'06)}, pages 459--468. IEEE, 2006.

\bibitem{andoni2014beyond}
Alexandr Andoni, Piotr Indyk, Huy~L Nguyen, and Ilya Razenshteyn.
\newblock Beyond locality-sensitive hashing.
\newblock In {\em Proceedings of the twenty-fifth annual ACM-SIAM symposium on
  Discrete algorithms}, pages 1018--1028. SIAM, 2014.

\bibitem{andoni2018approximate}
Alexandr Andoni, Piotr Indyk, and Ilya Razenshteyn.
\newblock Approximate nearest neighbor search in high dimensions.
\newblock In {\em Proceedings of the International Congress of Mathematicians:
  Rio de Janeiro 2018}, pages 3287--3318. World Scientific, 2018.

\bibitem{andoni2015optimal}
Alexandr Andoni and Ilya Razenshteyn.
\newblock Optimal data-dependent hashing for approximate near neighbors.
\newblock In {\em Proceedings of the forty-seventh annual ACM symposium on
  Theory of computing}, pages 793--801, 2015.

\bibitem{arandjelovic2013all}
Relja Arandjelovic and Andrew Zisserman.
\newblock All about vlad.
\newblock In {\em Proceedings of the IEEE conference on Computer Vision and
  Pattern Recognition}, pages 1578--1585, 2013.

\bibitem{arjovsky2017wasserstein}
Martin Arjovsky, Soumith Chintala, and L{\'e}on Bottou.
\newblock Wasserstein generative adversarial networks.
\newblock In {\em International conference on machine learning}, pages
  214--223. PMLR, 2017.

\bibitem{balaji2019normalized}
Yogesh Balaji, Rama Chellappa, and Soheil Feizi.
\newblock Normalized wasserstein for mixture distributions with applications in
  adversarial learning and domain adaptation.
\newblock In {\em Proceedings of the IEEE/CVF International Conference on
  Computer Vision}, pages 6500--6508, 2019.

\bibitem{bentley1975multidimensional}
Jon~Louis Bentley.
\newblock Multidimensional binary search trees used for associative searching.
\newblock {\em Communications of the ACM}, 18(9):509--517, 1975.

\bibitem{boiman2008defense}
Oren Boiman, Eli Shechtman, and Michal Irani.
\newblock In defense of nearest-neighbor based image classification.
\newblock In {\em 2008 IEEE Conference on Computer Vision and Pattern
  Recognition}, pages 1--8. IEEE, 2008.

\bibitem{bonneel2015sliced}
Nicolas Bonneel, Julien Rabin, Gabriel Peyr{\'e}, and Hanspeter Pfister.
\newblock Sliced and {R}adon {W}asserstein barycenters of measures.
\newblock {\em Journal of Mathematical Imaging and Vision}, 51(1):22--45, 2015.

\bibitem{bonnotte2013unidimensional}
Nicolas Bonnotte.
\newblock {\em Unidimensional and evolution methods for optimal
  transportation}.
\newblock PhD thesis, Universit\'e Paris 11, France, 2013.

\bibitem{charikar2002similarity}
Moses~S Charikar.
\newblock Similarity estimation techniques from rounding algorithms.
\newblock In {\em Proceedings of the thiry-fourth annual ACM symposium on
  Theory of computing}, pages 380--388, 2002.

\bibitem{courty2016optimal}
Nicolas Courty, R{\'e}mi Flamary, Devis Tuia, and Alain Rakotomamonjy.
\newblock Optimal transport for domain adaptation.
\newblock {\em IEEE transactions on pattern analysis and machine intelligence},
  39(9):1853--1865, 2016.

\bibitem{damodaran2018deepjdot}
Bharath~Bhushan Damodaran, Benjamin Kellenberger, R{\'e}mi Flamary, Devis Tuia,
  and Nicolas Courty.
\newblock Deepjdot: Deep joint distribution optimal transport for unsupervised
  domain adaptation.
\newblock In {\em Proceedings of the European Conference on Computer Vision
  (ECCV)}, pages 447--463, 2018.

\bibitem{datar2004locality}
Mayur Datar, Nicole Immorlica, Piotr Indyk, and Vahab~S Mirrokni.
\newblock Locality-sensitive hashing scheme based on p-stable distributions.
\newblock In {\em Proceedings of the twentieth annual symposium on
  Computational geometry}, pages 253--262, 2004.

\bibitem{pointCloudMNIST}
Cristian Garcia.
\newblock Point cloud mnist 2d, 2020.

\bibitem{gretton2006kernel}
Arthur Gretton, Karsten Borgwardt, Malte Rasch, Bernhard Sch{\"o}lkopf, and
  Alex Smola.
\newblock A kernel method for the two-sample-problem.
\newblock {\em Advances in neural information processing systems}, 19:513--520,
  2006.

\bibitem{gulrajani2017improved}
Ishaan Gulrajani, Faruk Ahmed, Martin Arjovsky, Vincent Dumoulin, and Aaron~C
  Courville.
\newblock {Improved training of Wasserstein GANs}.
\newblock In {\em Advances in Neural Information Processing Systems}, pages
  5767--5777, 2017.

\bibitem{indyk1998approximate}
Piotr Indyk and Rajeev Motwani.
\newblock Approximate nearest neighbors: towards removing the curse of
  dimensionality.
\newblock In {\em Proceedings of the thirtieth annual ACM symposium on Theory
  of computing}, pages 604--613, 1998.

\bibitem{jebara2004probability}
Tony Jebara, Risi Kondor, and Andrew Howard.
\newblock Probability product kernels.
\newblock {\em The Journal of Machine Learning Research}, 5:819--844, 2004.

\bibitem{jegou2010aggregating}
Herv{\'e} J{\'e}gou, Matthijs Douze, Cordelia Schmid, and Patrick P{\'e}rez.
\newblock Aggregating local descriptors into a compact image representation.
\newblock In {\em 2010 IEEE computer society conference on computer vision and
  pattern recognition}, pages 3304--3311. IEEE, 2010.

\bibitem{JDH17}
Jeff Johnson, Matthijs Douze, and Herv{\'e} J{\'e}gou.
\newblock Billion-scale similarity search with gpus.
\newblock {\em arXiv preprint arXiv:1702.08734}, 2017.

\bibitem{kalantidis2016cross}
Yannis Kalantidis, Clayton Mellina, and Simon Osindero.
\newblock Cross-dimensional weighting for aggregated deep convolutional
  features.
\newblock In {\em European conference on computer vision}, pages 685--701.
  Springer, 2016.

\bibitem{kaplan2020locality}
Haim Kaplan and Jay Tenenbaum.
\newblock Locality sensitive hashing for set-queries, motivated by group
  recommendations.
\newblock In {\em 17th Scandinavian Symposium and Workshops on Algorithm Theory
  (SWAT 2020)}. Schloss Dagstuhl-Leibniz-Zentrum f{\"u}r Informatik, 2020.

\bibitem{kolouri2021wasserstein}
Soheil Kolouri, Navid Naderializadeh, Gustavo~K. Rohde, and Heiko Hoffmann.
\newblock Wasserstein embedding for graph learning.
\newblock In {\em International Conference on Learning Representations}, 2021.

\bibitem{kolouri2019generalized}
Soheil Kolouri, Kimia Nadjahi, Umut Simsekli, Roland Badeau, and Gustavo Rohde.
\newblock Generalized sliced wasserstein distances.
\newblock In {\em Advances in Neural Information Processing Systems}, pages
  261--272, 2019.

\bibitem{kolouri2016radon}
Soheil Kolouri, Se~Rim Park, and Gustavo~K. Rohde.
\newblock The {R}adon cumulative distribution transform and its application to
  image classification.
\newblock {\em Image Processing, IEEE Transactions on}, 25(2):920--934, 2016.

\bibitem{kolouri2017optimal}
Soheil Kolouri, Se~Rim Park, Matthew Thorpe, Dejan Slepcev, and Gustavo~K
  Rohde.
\newblock Optimal mass transport: Signal processing and machine-learning
  applications.
\newblock {\em IEEE Signal Processing Magazine}, 34(4):43--59, 2017.

\bibitem{kolouri2018sliced}
Soheil Kolouri, Phillip~E. Pope, Charles~E. Martin, and Gustavo~K. Rohde.
\newblock Sliced {W}asserstein auto-encoders.
\newblock In {\em International Conference on Learning Representations}, 2019.

\bibitem{kolouri2016sliced}
Soheil Kolouri, Yang Zou, and Gustavo~K Rohde.
\newblock Sliced-{W}asserstein kernels for probability distributions.
\newblock In {\em Proceedings of the IEEE Conference on Computer Vision and
  Pattern Recognition}, pages 4876--4884, 2016.

\bibitem{kulis2009kernelized}
Brian Kulis and Kristen Grauman.
\newblock Kernelized locality-sensitive hashing for scalable image search.
\newblock In {\em 2009 IEEE 12th international conference on computer vision},
  pages 2130--2137. IEEE, 2009.

\bibitem{lecun1998gradient}
Yann LeCun, L{\'e}on Bottou, Yoshua Bengio, and Patrick Haffner.
\newblock Gradient-based learning applied to document recognition.
\newblock {\em Proceedings of the IEEE}, 86(11):2278--2324, 1998.

\bibitem{lee2019sliced}
Chen-Yu Lee, Tanmay Batra, Mohammad~Haris Baig, and Daniel Ulbricht.
\newblock Sliced wasserstein discrepancy for unsupervised domain adaptation.
\newblock In {\em Proceedings of the IEEE/CVF Conference on Computer Vision and
  Pattern Recognition}, pages 10285--10295, 2019.

\bibitem{lee2019set}
Juho Lee, Yoonho Lee, Jungtaek Kim, Adam Kosiorek, Seungjin Choi, and Yee~Whye
  Teh.
\newblock Set transformer: A framework for attention-based
  permutation-invariant neural networks.
\newblock In {\em International Conference on Machine Learning}, pages
  3744--3753. PMLR, 2019.

\bibitem{levine2020wasserstein}
Alexander Levine and Soheil Feizi.
\newblock Wasserstein smoothing: Certified robustness against wasserstein
  adversarial attacks.
\newblock In {\em International Conference on Artificial Intelligence and
  Statistics}, pages 3938--3947. PMLR, 2020.

\bibitem{liutkus2019sliced}
Antoine Liutkus, Umut Simsekli, Szymon Majewski, Alain Durmus, and
  Fabian-Robert St{\"o}ter.
\newblock Sliced-wasserstein flows: Nonparametric generative modeling via
  optimal transport and diffusions.
\newblock In {\em International Conference on Machine Learning}, pages
  4104--4113. PMLR, 2019.

\bibitem{csimcsekli2018sliced}
Antoine Liutkus, Umut {\c{S}}im{\c{s}}ekli, Szymon Majewski, Alain Durmus, and
  Fabian-Robert Stoter.
\newblock Sliced-{W}asserstein flows: Nonparametric generative modeling via
  optimal transport and diffusions.
\newblock In {\em International Conference on Machine Learning}, 2019.

\bibitem{mialon2021a}
Gr{\'e}goire Mialon, Dexiong Chen, Alexandre d'Aspremont, and Julien Mairal.
\newblock A trainable optimal transport embedding for feature aggregation and
  its relationship to attention.
\newblock In {\em International Conference on Learning Representations}, 2021.

\bibitem{moosmuller2020linear}
Caroline Moosm{\"u}ller and Alexander Cloninger.
\newblock Linear optimal transport embedding: Provable fast wasserstein
  distance computation and classification for nonlinear problems.
\newblock {\em arXiv preprint arXiv:2008.09165}, 2020.

\bibitem{muandet2012learning}
Krikamol Muandet, Kenji Fukumizu, Francesco Dinuzzo, and Bernhard
  Sch{\"o}lkopf.
\newblock Learning from distributions via support measure machines.
\newblock In {\em Proceedings of the 25th International Conference on Neural
  Information Processing Systems-Volume 1}, pages 10--18, 2012.

\bibitem{murphy2018janossy}
Ryan~L. Murphy, Balasubramaniam Srinivasan, Vinayak Rao, and Bruno Ribeiro.
\newblock Janossy pooling: Learning deep permutation-invariant functions for
  variable-size inputs.
\newblock In {\em International Conference on Learning Representations}, 2019.

\bibitem{nadjahi2020statistical}
Kimia Nadjahi, Alain Durmus, L{\'e}na{\"\i}c Chizat, Soheil Kolouri, Shahin
  Shahrampour, and Umut {\c{S}}im{\c{s}}ekli.
\newblock Statistical and topological properties of sliced probability
  divergences.
\newblock In {\em Advances in Neural Information Processing Systems}, 2020.

\bibitem{nagarkar2018pslsh}
Parth Nagarkar and K~Sel{\c{c}}uk Candan.
\newblock Pslsh: An index structure for efficient execution of set queries in
  high-dimensional spaces.
\newblock In {\em Proceedings of the 27th ACM International Conference on
  Information and Knowledge Management}, pages 477--486, 2018.

\bibitem{park2018cumulative}
Se~Rim Park, Soheil Kolouri, Shinjini Kundu, and Gustavo~K Rohde.
\newblock The cumulative distribution transform and linear pattern
  classification.
\newblock {\em Applied and Computational Harmonic Analysis}, 45(3):616--641,
  2018.

\bibitem{paty2019subspace}
Fran{\c{c}}ois-Pierre Paty and Marco Cuturi.
\newblock Subspace robust wasserstein distances.
\newblock In {\em International Conference on Machine Learning}, 2019.

\bibitem{peyre2018computational}
Gabriel Peyr{\'e} and Marco Cuturi.
\newblock Computational optimal transport.
\newblock {\em arXiv preprint arXiv:1803.00567}, 2018.

\bibitem{philbin2007object}
James Philbin, Ondrej Chum, Michael Isard, Josef Sivic, and Andrew Zisserman.
\newblock Object retrieval with large vocabularies and fast spatial matching.
\newblock In {\em 2007 IEEE conference on computer vision and pattern
  recognition}, pages 1--8. IEEE, 2007.

\bibitem{poczos2012nonparametric}
Barnab{\'a}s P{\'o}czos and Jeff Schneider.
\newblock Nonparametric estimation of conditional information and divergences.
\newblock In {\em Artificial Intelligence and Statistics}, pages 914--923.
  PMLR, 2012.

\bibitem{poczos2011nonparametric}
Barnab{\'a}s P{\'o}czos, Liang Xiong, and Jeff Schneider.
\newblock Nonparametric divergence estimation with applications to machine
  learning on distributions.
\newblock In {\em Proceedings of the Twenty-Seventh Conference on Uncertainty
  in Artificial Intelligence}, pages 599--608, 2011.

\bibitem{rabin2012wasserstein}
Julien Rabin, Gabriel Peyr{\'e}, Julie Delon, and Marc Bernot.
\newblock {W}asserstein barycenter and its application to texture mixing.
\newblock In {\em Scale Space and Variational Methods in Computer Vision},
  pages 435--446. Springer, 2012.

\bibitem{radenovic2018fine}
Filip Radenovi{\'c}, Giorgos Tolias, and Ond{\v{r}}ej Chum.
\newblock Fine-tuning cnn image retrieval with no human annotation.
\newblock {\em IEEE transactions on pattern analysis and machine intelligence},
  41(7):1655--1668, 2018.

\bibitem{ILSVRC15}
Olga Russakovsky, Jia Deng, Hao Su, Jonathan Krause, Sanjeev Satheesh, Sean Ma,
  Zhiheng Huang, Andrej Karpathy, Aditya Khosla, Michael Bernstein,
  Alexander~C. Berg, and Li Fei-Fei.
\newblock {ImageNet Large Scale Visual Recognition Challenge}.
\newblock {\em International Journal of Computer Vision (IJCV)},
  115(3):211--252, 2015.

\bibitem{shakhnarovich2008nearest}
Gregory Shakhnarovich, Trevor Darrell, and Piotr Indyk.
\newblock Nearest-neighbor methods in learning and vision.
\newblock {\em IEEE Trans. Neural Networks}, 19(2):377, 2008.

\bibitem{shifat2021radon}
Mohammad Shifat-E-Rabbi, Xuwang Yin, Abu Hasnat~Mohammad Rubaiyat, Shiying Li,
  Soheil Kolouri, Akram Aldroubi, Jonathan~M Nichols, and Gustavo~K Rohde.
\newblock Radon cumulative distribution transform subspace modeling for image
  classification.
\newblock {\em Journal of Mathematical Imaging and Vision}, 63(9):1185--1203,
  2021.

\bibitem{simonyan2014very}
Karen Simonyan and Andrew Zisserman.
\newblock Very deep convolutional networks for large-scale image recognition.
\newblock {\em arXiv preprint arXiv:1409.1556}, 2014.

\bibitem{sinha2018certifying}
Aman Sinha, Hongseok Namkoong, and John Duchi.
\newblock Certifying some distributional robustness with principled adversarial
  training.
\newblock In {\em International Conference on Learning Representations}, 2018.

\bibitem{togninalli2019wasserstein}
Matteo Togninalli, Elisabetta Ghisu, Felipe Llinares-L{\'o}pez, Bastian Rieck,
  and Karsten Borgwardt.
\newblock Wasserstein weisfeiler-lehman graph kernels.
\newblock {\em Advances in Neural Information Processing Systems},
  32:6439--6449, 2019.

\bibitem{tolstikhin2018wasserstein}
Ilya Tolstikhin, Olivier Bousquet, Sylvain Gelly, and Bernhard Schoelkopf.
\newblock Wasserstein auto-encoders.
\newblock In {\em International Conference on Learning Representations}, 2018.

\bibitem{villani2008optimal}
C{\'e}dric Villani.
\newblock {\em Optimal transport: old and new}, volume 338.
\newblock Springer Science \& Business Media, 2008.

\bibitem{wagstaff2019limitations}
Edward Wagstaff, Fabian Fuchs, Martin Engelcke, Ingmar Posner, and Michael~A
  Osborne.
\newblock On the limitations of representing functions on sets.
\newblock In {\em International Conference on Machine Learning}, pages
  6487--6494. PMLR, 2019.

\bibitem{wald2006building}
Ingo Wald and Vlastimil Havran.
\newblock On building fast kd-trees for ray tracing, and on doing that in o (n
  log n).
\newblock In {\em 2006 IEEE Symposium on Interactive Ray Tracing}, pages
  61--69. IEEE, 2006.

\bibitem{wang2020deep}
Qilong Wang, Jiangtao Xie, Wangmeng Zuo, Lei Zhang, and Peihua Li.
\newblock Deep cnns meet global covariance pooling: Better representation and
  generalization.
\newblock {\em IEEE transactions on pattern analysis and machine intelligence},
  2020.

\bibitem{wang2013linear}
Wei Wang, Dejan Slep{\v{c}}ev, Saurav Basu, John~A Ozolek, and Gustavo~K Rohde.
\newblock A linear optimal transportation framework for quantifying and
  visualizing variations in sets of images.
\newblock {\em International journal of computer vision}, 101(2):254--269,
  2013.

\bibitem{wong2019wasserstein}
Eric Wong, Frank Schmidt, and Zico Kolter.
\newblock Wasserstein adversarial examples via projected sinkhorn iterations.
\newblock In {\em International Conference on Machine Learning}, pages
  6808--6817. PMLR, 2019.

\bibitem{wu2020stronger}
Kaiwen Wu, Allen Wang, and Yaoliang Yu.
\newblock Stronger and faster wasserstein adversarial attacks.
\newblock In {\em International Conference on Machine Learning}, pages
  10377--10387. PMLR, 2020.

\bibitem{wu20153d}
Zhirong Wu, Shuran Song, Aditya Khosla, Fisher Yu, Linguang Zhang, Xiaoou Tang,
  and Jianxiong Xiao.
\newblock 3d shapenets: A deep representation for volumetric shapes.
\newblock In {\em Proceedings of the IEEE conference on computer vision and
  pattern recognition}, pages 1912--1920, 2015.

\bibitem{xiong2014learning}
Liang Xiong and Jeff Schneider.
\newblock Learning from point sets with observational bias.
\newblock In {\em Proceedings of the Thirtieth Conference on Uncertainty in
  Artificial Intelligence}, pages 898--906, 2014.

\bibitem{zaheer2017deep}
Manzil Zaheer, Satwik Kottur, Siamak Ravanbhakhsh, Barnab{\'a}s P{\'o}czos,
  Ruslan Salakhutdinov, and Alexander~J Smola.
\newblock Deep sets.
\newblock In {\em Proceedings of the 31st International Conference on Neural
  Information Processing Systems}, pages 3394--3404, 2017.

\bibitem{zhang2019fspool}
Yan Zhang, Jonathon Hare, and Adam Pr{\"u}gel-Bennett.
\newblock Fspool: Learning set representations with featurewise sort pooling.
\newblock {\em arXiv preprint arXiv:1906.02795}, 2019.

\bibitem{Zhang2020FSPool}
Yan Zhang, Jonathon Hare, and Adam Prügel-Bennett.
\newblock Fspool: Learning set representations with featurewise sort pooling.
\newblock In {\em International Conference on Learning Representations}, 2020.

\end{thebibliography}
}

\section{Supplementary Materials}

\subsection{Proofs}

Here we include the proof for the C1 and C2 conditions covered in Section 4. Recall that $\mu_i$ represents a probability measure, $\mu_i^\theta$ represents $g_{\theta\#}\mu_i$, where $g_\theta:\mathbb{R}^d\rightarrow \mathbb{R}$ with some regularity constraints, and that we define the cumulative distribution transform (CDT) \cite{park2018cumulative} of $\mu_i^\theta$ as
\begin{align*}\label{eq:cdt}
\phi_\theta(\mu_i)\coloneqq (T_i^\theta - id),
\end{align*}
where $T_i^\theta$ is the Monge map/coupling, and $id$ denotes the identity function. For a fixed $\theta$, here we prove that $\phi^\theta(\mu_i)$ satisfies the following conditions:

\begin{table}[t!]
    \centering
    \begin{tabular}{c|c}
       Method  &  Complexity \\
       \hline
        Gem-p   &  $\mathcal{O}(Np^2 d)$\\
        Cov.    &  $\mathcal{O}(Nd^2)$\\
        FSPool  &  $\mathcal{O}(Nd\text{log}N)$\\
        \hline
        SLOSH  & $\mathcal{O}(LN(d+\text{log}N))$
    \end{tabular}
    \caption{Computational complexities.}
    \label{tab:my_label}
\end{table}

{\bf C1.}  The weighted $\ell_2$-norm of the embedded slice, $\phi_\theta(\mu_i)$, satisfies:
    \begin{align*}
        \|\phi_\theta(\mu_i)\|_{\mu_0^\theta,2}&=\left(\int_{\mathbb{R}}\|\phi_\theta(\mu_i(t))\|_2^2 d\mu_0^\theta(t)\right)^{\frac{1}{2}} \\&=\mathcal{W}_2(\mu_i^\theta,\mu_0^\theta),
    \end{align*}
    
\begin{proof}
    We start by writing the squared distance:
    \begin{align*}
    &\|\phi_\theta(\mu_i)\|^2_{\mu_0^\theta,2}=\int_{\mathbb{R}}\|\phi_\theta(\mu_i(t))\|_2^2 d\mu_0^\theta(t)\\
    &=\int_{\mathbb{R}}\|T_i^\theta(t)-t\|_2^2 d\mu_0^\theta(t)&\\
        &=\int_{\mathbb{R}}\| (F_{\mu_i^\theta}^{-1}\circ F_{\mu_0^\theta})(t)-t\|_2^2 d\mu_0^\theta(t)&\\
        &=\int^{1}_{0}\|F_{\mu_i^\theta}^{-1}(\tau)-F_{\mu_0^\theta}^{-1}(\tau)\|_2^2 d\tau&\\
        &= \mathcal{W}_2^2(\mu_i^\theta,\mu_0^\theta)&
    \end{align*}
where we used the definition of the one-dimensional Monge map, $T^\theta_i = F_{\mu_i^\theta}^{-1}\circ F_{\mu_0^\theta}$, and the change of variable $\tau=F_{\mu_0^\theta}(t)$. The corollary, $\|\phi_\theta(\mu_0)\|_{\mu_0^\theta,2}=0$, is trivial as $\mathcal{W}_2^2(\mu_0^\theta,\mu_0^\theta)=0$.
    \end{proof}

{\bf C2.} The weighted $\ell_2$ distance satisfies:
    \begin{align}
        \|\phi_\theta(\mu_i)-\phi_\theta(\mu_j)\|_{\mu_0^\theta,2}=\mathcal{W}_2(\mu_i^\theta,\mu_j^\theta).       
    \end{align}
    \begin{proof}
    Similar to the previous proof:
    \begin{align*}
    &\|\phi_\theta(\mu_i)-\phi_\theta(\mu_j)\|^2_{\mu_0^\theta,2}=\int_{\mathbb{R}}\|\phi_\theta(\mu_i(t))-\phi_\theta(\mu_j(t))\|_2^2 d\mu_0^\theta(t)\\
    &=\int_{\mathbb{R}}\|T_i^\theta(t)-T_j^\theta(t)\|_2^2 d\mu_0^\theta(t)&\\
    &=\int_{\mathbb{R}}\| (F_{\mu_i^\theta}^{-1}\circ F_{\mu_0^\theta})(t)-(F_{\mu_j^\theta}^{-1}\circ F_{\mu_0^\theta})(t)\|_2^2 d\mu_0^\theta(t)&\\
        &=\int^{1}_{0}\|F_{\mu_i^\theta}^{-1}(\tau)-F_{\mu_j^\theta}^{-1}(\tau)\|_2^2 d\tau
        = \mathcal{W}_2^2(\mu_i^\theta,\mu_j^\theta)&
    \end{align*}
    where again we used the definition of the one-dimensional Monge map, $T^\theta_i = F_{\mu_i^\theta}^{-1}\circ F_{\mu_0^\theta}$, and the change of variable $\tau=F_{\mu_0^\theta}(t)$. 
    \end{proof}

As a corollary of C1 and C2 we have: 
\begin{align*}
&\mathcal{GSW}^2_2(\mu_i,\mu_j)=\int_{\Omega_\theta} \mathcal{W}_2^2(\mu_i^\theta,\mu_j^\theta) d\sigma(\theta)=\\
&\int_{\Omega_\theta}\|\phi_\theta(\mu_i)-\phi_\theta(\mu_j)\|_{\mu_0^\theta, 2}^2 d\sigma(\theta)
\end{align*}

\subsection{Computational complexities}

For the sake of completeness, here we include the computational complexities of the baseline methods used in our paper. In Table \ref{tab:my_label},  we provide the computational complexity of embedding a set $X=\{x_n\in\mathbb{R}^d\}_{n=1}^N$ into a vector space.

\end{document}